\documentclass{article} % For LaTeX2e
\usepackage{iclr2026_conference,times}
\usepackage{comment}
\usepackage{algorithm}
\usepackage{algpseudocode}
\usepackage{amssymb}
\usepackage{graphicx}
\usepackage{placeins}

\usepackage{amsmath,amsfonts,bm}

\def\eqref#1{equation~\ref{#1}}
\def\1{\bm{1}}

\DeclareMathAlphabet{\mathsfit}{\encodingdefault}{\sfdefault}{m}{sl}
\SetMathAlphabet{\mathsfit}{bold}{\encodingdefault}{\sfdefault}{bx}{n}

\usepackage{hyperref}
\usepackage{url}
\usepackage{booktabs}

\iclrfinalcopy

\title{How Clued up are LLMs? Evaluating Multi-Step Deductive Reasoning in a Text-Based Game Environment}

\author{
Rebecca Ansell \\
Department of Computer Science \\
Georgetown University \\
Washington, DC, USA \\
\texttt{rja80@georgetown.edu}
\And
Autumn Toney-Wails \\
Syntheos, Corp \\
UNU-Merit \\
Maastricht, NL \\
\texttt{toney@merit.unu.edu}
}

\begin{document}

\maketitle

\begin{abstract}
Deducing \textit{whodunit} proves challenging for LLM agents. In this paper, we implement a text-based multi-agent version of the classic board game Clue as a rule-based testbed for evaluating multi-step deductive reasoning, with six agents drawn from GPT-4o-mini and Gemini-2.5-Flash. We further investigate whether fine-tuning on structured logic puzzles transfers to improved in-game reasoning and gameplay. Across 18 simulated games, agents achieve only four correct wins, indicating difficulty in maintaining consistent deductive reasoning over the course of a full game. Additionally, we find that fine-tuning does not reliably improve performance and, in some cases, appears to increase reasoning volume without improving reasoning precision.

%We find that agentic Clue players struggle to perform consistent deductive reasoning, resulting in only four wins across 18 simulated games, and that fine-tuning does not reliably improve deductive performance, and in some cases encourages reasoning volume over reasoning precision. %---the model that makes the most deductions achieves the worst accusation accuracy. 
\end{abstract}

\section{Introduction}

%--------------CLEAN TEXT
Recent improvements in the performance of large language models (LLMs) on tasks extending beyond traditional natural language processing (e.g., summarization and translation) have enabled their use as general-purpose agents in interactive environments \cite{bandi2025rise}. Agentic deployments of LLMs require structured reasoning capabilities, including deductive, strategic, and temporally extended inference, differing from single-turn query–response settings \cite{wu2024autogen, sypherd2024practical,nisa2025agentic,zhou2025exploring}. In agentic settings, response accuracy alone is insufficient; understanding how an agent arrives at a conclusion, determining if it applied the appropriate form of reasoning, and evaluating the logical soundness of its intermediate reasoning steps are equally critical \cite{xia2025evaluating,wang2025dyflow, liu2026gem}. 

Different scenarios require distinct forms of reasoning. For example, algebraic word problems and logic puzzles are commonly used to probe specific reasoning processes. Algebraic word problems typically require multi-step deductive reasoning and the translation of natural language into formal representations, whereas logic grid puzzles emphasize constraint satisfaction and systematic elimination. Prior work has focused on designing various environments to evaluate LLM reasoning across a wide range of real-world simulated tasks. One such approach is to evaluate LLMs acting as autonomous agents within interactive game environments, where they must reason dynamically under structured rules and evolving state conditions \cite{zhang-etal-2024-probing,chi2024amongagentsevaluatinglargelanguage,xu2025languageagentsreinforcementlearning}. In this work, we focus on evaluating deductive reasoning in this game setting. 

We design an experiment inspired by the classic murder mystery board game Clue, a deduction-based game where players must identify a hidden solution by integrating evidence gathered across multiple turns. Clue provides a natural framework for evaluating agentic deductive reasoning capabilities because game play is structured as an information-constrained logic puzzle. During game play in Clue, partial information is revealed indirectly and players must make corresponding inferences through logical elimination rather than explicit confirmation. Game success depends on a player's ability to correctly apply logical rules to constrain what remains possible while strategically managing the information revealed to other players. This setting allows us to test whether LLM agents can sustain multi-step, logically sound reasoning over extended interactions without introducing inconsistent or invalid inferences.

We implement a modified version of the board game by designing a text-based Clue environment with GPT-4o-mini \cite{openai2024gpt4ocard} and Gemini-2.5-Flash \cite{comanici2025gemini} as player agents. Additionally, we include a fine-tuning experiment to investigate whether targeted exposure to related deductive reasoning tasks can improve agentic game play in the Clue environment. We fine-tune each LLM on Mind Bender logic puzzles, which include a word-problem statement, a set of constraints, and a corresponding solution with step-by-step deductive reasoning explanations \cite{criticalthinking_mindbenders}.

Taking the logs of 18 simulated games, we conduct a comparative evaluation of GPT-4o-mini and Gemini-2.5-Flash using their baseline and fine-tuned model variants. Across our experiments we aim to answer three main research questions: 

\begin{itemize}
    \item \textbf{RQ1} Can LLM agents sustain logically consistent deductive reasoning across extended, multi-turn interactions in a structured game environment?
    \item \textbf{RQ2} Does fine-tuning on deductive reasoning tasks improve agentic performance in a related environment? 
    \item \textbf{RQ3} How does knowledge accumulation relate to reasoning quality in fine-tuned versus base model variants?
\end{itemize}

Our findings suggest that (1) LLMs struggle to successfully use valid deductive reasoning to achieve accurate outcomes in a game environment; (2) fine-tuning models on related logic puzzles does not reliably improve deductive performance and, in some cases, degrades it; and (3) there is a disconnect between information accumulation and reasoning quality as the model that accumulates the most knowledge and produces the most deductions performs worst, while the model that deduces least performs best, suggesting that reasoning precision matters more than reasoning volume in this setting.

\section{Related Work}

\subsection{Agentic Reasoning}

% Deductive reasoning and logical consistency in LLMs
LLMs have demonstrated improved performance on a range of reasoning benchmarks, particularly through Chain-of-Thought (CoT) prompting, which encourages models to generate intermediate reasoning steps before producing a final answer. Survey work highlights consistent gains on mathematical, commonsense, and symbolic reasoning tasks under CoT-style prompting \cite{chu-etal-2024-navigate}. However, subsequent studies show that these gains do not necessarily translate into robust multi-step reasoning. Empirical evaluations reveal performance degradation as reasoning depth increases and the presence of logical inconsistencies in generated reasoning chains \cite{patel2024multilogievalevaluatingmultisteplogical}. Moreover, recent work demonstrates that models can produce correct answers despite flawed intermediate reasoning, suggesting that generated explanations may not faithfully reflect the process underlying the model’s final prediction \cite{xu2026correctnessrewardingfaithfulreasoning, zheng2025cursecotlimitationschainofthought}.

% benchmarks for logical reasoning & multi step reasoning, state tracking 
Standard benchmarks for evaluating logical reasoning reflect this limitation. Datasets spanning mathematical problem solving (Mr-GSM8K \cite{zeng2024mrgsm8kmetareasoningbenchmarklarge}) and broader mathematical reasoning tasks \cite{hendrycks2021measuringmathematicalproblemsolving}, commonsense reasoning (CommonsenseQA \cite{talmor-etal-2019-commonsenseqa}), and logical inference (LogiQA 2.0 \cite{10174688}, ReClor \cite{yu2020reclorreadingcomprehensiondataset}) are broadly used to probe these capabilities, but they share a common structural constraint: each problem is presented in isolation, requiring a single inference or a short chain of inferences rather than sustained reasoning over an evolving knowledge state. As a result, they do not capture whether models can maintain logical consistency across dozens of sequential, interdependent inferences---a dimension of reasoning that becomes critical in long-horizon interactive settings.

\subsection{Agentic Game Play}

LLMs have been increasingly applied as agents in various game environments as a way to showcase and understand their capabilities in reasoning, negotiation, and communication. A growing body of work is evaluating LLMs in multi-agent games that involve complex decision-making and inference, suggesting that these models can serve as more than just text generators, but as interactive, strategic agents within a structured domain.

Game environments used to evaluate LLMs span several categories. Social deduction games like Werewolf and Avalon serve as benchmarks for evaluating LLMs' abilities in environments that demand deception, persuasion, and strategic planning. \cite{xu2025languageagentsreinforcementlearning} introduced a framework where LLM-based agents utilize reinforcement learning to enhance strategic play in Werewolf. The AmongAgents benchmark \cite{chi2024amongagentsevaluatinglargelanguage} has extended this line of work to a text-based analogue of \textit{Among Us}, where LLM agents must collaborate, deceive, and reason about social dynamics while completing tasks in a partially observable environment. 

Beyond social deduction, LLMs have been applied to negotiation and coalition-formation games, most notably in \textit{Diplomacy}. Meta's CICERO System \cite{doi:10.1126/science.ade9097} combined an LLM with a strategic reasoning module and achieved human-level performance in online Diplomacy, highlighting the potential for hybrid architectures that pair language models with explicit planning or game-theory components. Other work has explored LLM performance in games that require complex strategic or probabilistic reasoning---for example, poker, where models must reason about risk, uncertainty, and incomplete information \cite{zhuang2025pokerbenchtraininglargelanguage},  as well as a broader suite of board and video games designed to probe multi-step reasoning skills \cite{lin-etal-2025-gamebot}.

Existing game benchmarks, while often multi-turn, do not isolate deductive reasoning as the primary capability under evaluation. Social deduction games test intent modeling, trust, and persuasion rather than the accumulation of logical deductions over time. Strategic games demand probabilistic rather than deductive reasoning, requiring agents to weigh likelihoods rather than propagate hard constraints across a growing evidence set. As a result, it remains an open question whether LLMs can maintain a logically consistent belief state across sequential observations through deductive reasoning alone.

Our study addresses these research gaps by implementing a Clue-based game environment,  where agents must maintain and iteratively update a consistent belief state across sequential turns while integrating private information with publicly observed game interactions and propagating constraints over time. Different from frameworks for evaluating deductive reasoning at a single point in time, Clue requires agents to: (1) narrow candidate hypotheses through iterative elimination, (2) maintain logical consistency across an extended interaction, (3) integrate asymmetric information sources, and (4) determine when sufficient evidence has accumulated to make a confident accusation. By isolating long-horizon, stateful deduction in a controlled and non-social environment, Clue provides a testbed for evaluating LLM reasoning capabilities.

\section{The Clue Environment}

Clue (marketed as Cluedo outside of North America) is a murder mystery board game that requires strategy and deductive reasoning to win; a player must be the first to correctly identify the suspect, weapon, and location of the crime. For our experiments, we implement a variation of the game that does not require the board or dice. We describe the game setup in Section \ref{sec:setup} and the the formal task in Section \ref{sec:formal_task_def}.

\subsection{Game Setup and Rules}
\label{sec:setup}
The standard game includes 21 cards for each possible suspect, weapon, and room (location):
\begin{itemize}
    \item 6 suspects: Miss Scarlet, Colonel Mustard, Mrs. White, Mr. Green, Mrs. Peacock, and Professor Plum
    \item 6 weapons: Candlestick, Knife, Lead Pipe, Revolver, Rope, and Wrench
    \item 9 rooms: Kitchen, Ballroom, Conservatory, Dining Room, Billiard Room, Library, Lounge, Hall, and Study
\end{itemize}

At the start of the game, one card from each category is randomly selected and placed in a confidential envelope, forming the solution. The remaining cards are distributed among players. Players take turns making suggestions about the solution (e.g. ``I suggest it was Prof. Plum, in the Kitchen, with the Candlestick''). When a player makes a suggestion, other players are required to refute some component of the suggestion if they hold any matching cards (i.e., if a player holds the Prof. Plum card they must show the suggester). Only one card is shown to the suggester; however, if a player holds multiple matching cards, they can strategically choose which card to reveal. Additionally, the suggester is allowed to include a card she holds (i.e., if she holds the Candlestick card, her suggestion is valid), but her card remains secret to all other players. If no player can refute a suggestion, the suggester gains information that none of those cards are held by other players. 

\begin{figure}
    \centering
    \includegraphics[width=0.7\linewidth]{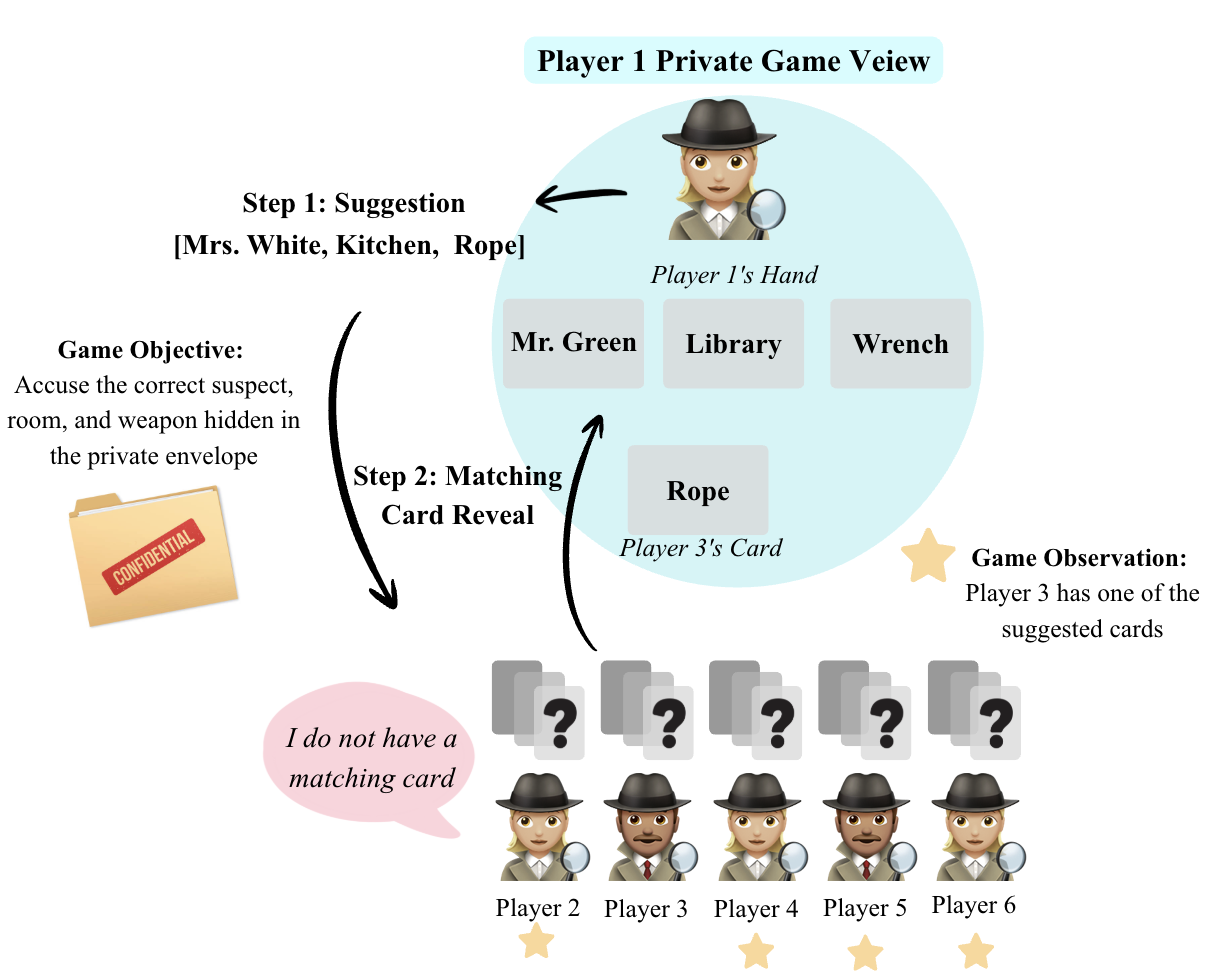}
    \caption{Clue gameplay diagram illustrating Player 1's turn. In this example, Player 1 made a suggestion (Mrs. White, Kitchen, Rope) and Player 2 did not have any matching cards to the suggestion, so then it moved to Player 3. Since Player 3 had a matching card, he revealed it to Player 1 (only). Player 1 can then rule out ``Rope'' as the correct weapon and Players 2, 4, 5, and 6 know that Player 3 must have Mrs. White, Kitchen, or Rope.}
    \label{fig:clue_diagram}
\end{figure}

A player’s turn ends after she makes a suggestion and all required cards are revealed (or none are revealed). Play then passes to the next player, and the game continues sequentially in this manner. Players may make suggestions consisting of any combination of suspect, weapon, and room, and suggestions may be repeated. Figure \ref{fig:clue_diagram} displays a diagram of the game.

To win the game, a player must make a formal accusation (in the same format as suggestions). If the accusation is correct, that player wins. If the accusation is incorrect, that player is eliminated and the game continues in the same way for the remaining players. The game ends when a correct accusation is made.  

\subsection{Formal Task Definition}
\label{sec:formal_task_def}
We formalize Clue as a multi-step deductive reasoning problem under partial observability. At each turn, an agent maintains a knowledge state, summarizing all the information acquired through the current turn:

\begin{itemize}
    \item The agent's private hand.
    \item The set of cards the agent has been shown by other players.
    \item The complete suggestion history: the suggester, the three cards suggested, and which player (if any) disproved it for each suggestion made.
    \item The agent's own reasoning from the previous turn for continuity.
\end{itemize}

Given the knowledge state, the agent must: 
(1) determine which cards remain possible in the envelope, (2) select a suggestion that maximally reduces uncertainty, and (3) decide when sufficient information has been gathered to make an accusation. We provide simplified pseudocode for the agentic implementation of the game in Algorithm \ref{alg:run_game} (See Appendix \ref{sec:code}), highlighting the phases that require prompting and player output.

\section{Experimental Design}

% \subsection{Agent Implementation}
We implement LLM-based agents that interact with the Clue environment through structured prompts and formatted responses. Our design separates game logic from agent reasoning. Here we provide details on the agentic player implementation (Sections [\ref{sec:aa} and \ref{sec:rp}]), LLM fine-tuning (Section [\ref{sec:ft}]), the experiment setup (Section [\ref{sec:es}]), and the evaluation metrics we use to analyze the results (Section [\ref{sec:em}]). 

\subsection{Agent Architecture}
\label{sec:aa}
On each turn, agents receive their knowledge state as defined in Section ~\ref{sec:formal_task_def} augmented with pre-computed derived information: remaining candidates per category, and locked candidates per category (if only one candidate is remaining), undisproved suggestions, and definitive answers. This design tests strategic reasoning with well-structured information rather than information from unstructured text. 

Agents respond in a structured format, ensuring parseable outputs and explicit reasoning:
\begin{verbatim}
SUMMARY: <1-2 sentence summary>
REASONING: <step-by-step deduction>
SUGGESTION: <suspect>, <weapon>, <room>
ACCUSATION: <suspect>, <weapon>, <room> or NONE
\end{verbatim}
Responses failing to parse or lacking reasoning are re-prompted up to 3 times before falling back to random action.

\subsection{Reasoning Phases}
\label{sec:rp}
There are two main reasoning phases in the game (See Algorithm ~\ref{alg:run_game}). \textbf{Phase 1 (Deduction):} Agents analyze game history to deduce opponent cards through cross-referencing disprovals. Deductions are classified as correct or incorrect to measure accuracy. The agents are prompted with clear rules and instructions on how to make deductions, along with the knowledge they have acquired up until that turn (See Appendix \ref{sec:prompt_wording}). \textbf{Phase 2 (Action):} Using updated knowledge, agents select suggestions and decide whether to accuse. Agents are prompted explicitly at this phase to make an optimal game action (See Appendix \ref{sec:prompt_wording}). 

There is also a third phase highlighted in Algorithm ~\ref{alg:run_game} that is mainly focused on strategic gameplay, but does require valid reasoning for optimal strategy. \textbf{Phase 3 (Show-card): } If an agent has a matching card to a component in a suggestion, they must show that card; however, if they have more than one matching card they are prompted to strategically select the card to reveal (See Appendix \ref{sec:prompt_wording}).

\subsection{Fine-tuning Details}
\label{sec:ft}
\begin{figure}
    \centering
    \includegraphics[width=0.92\linewidth]{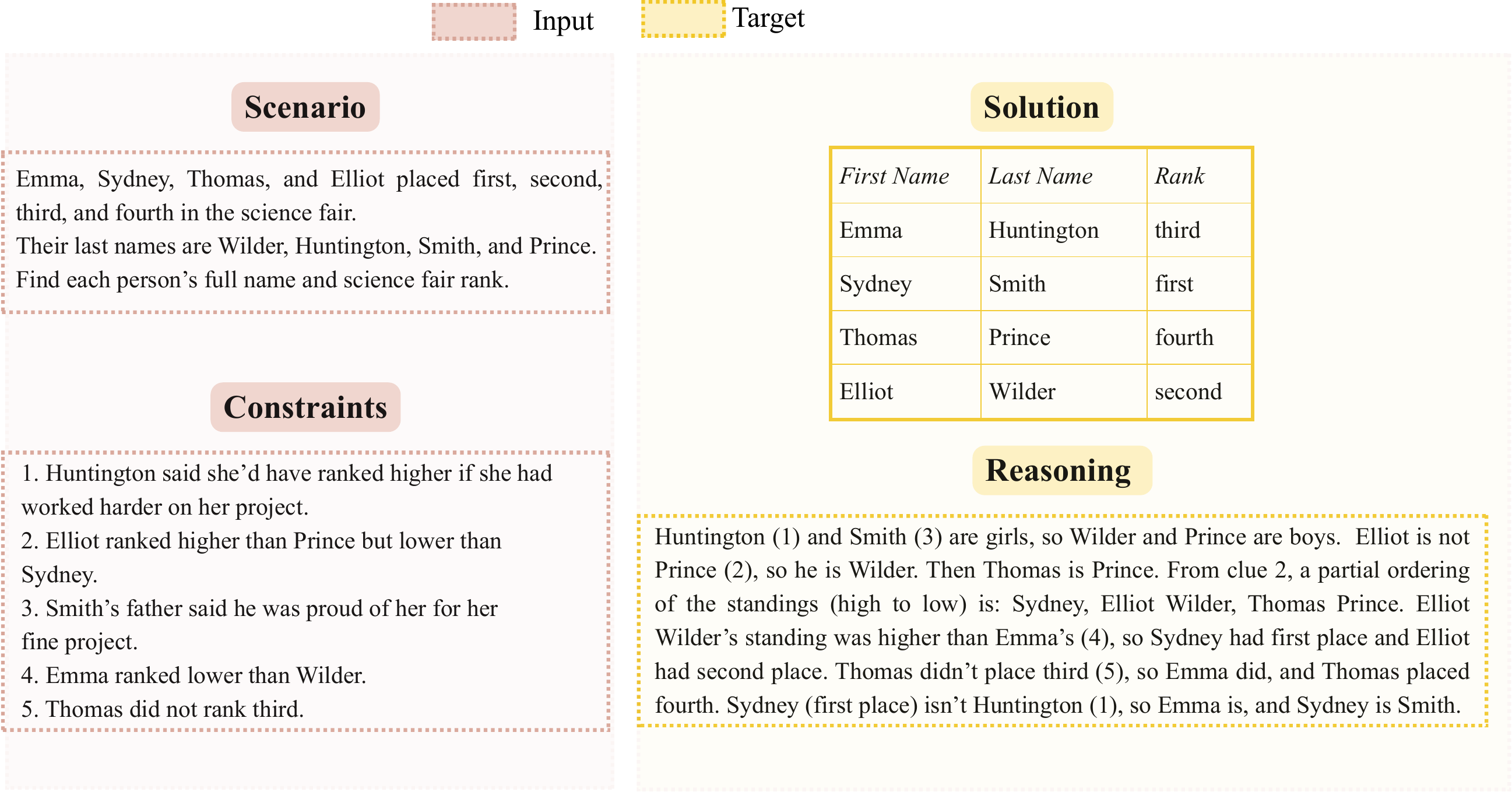}
    \caption{Mind Bender fine-tuning example (adopted from original).}
    \label{fig:mindbender}
\end{figure}

To train our fine-tuned model variants, we use a 50-sample training dataset and a 10-sample validation dataset of problems and solutions drawn from the Mind Bender puzzle series \cite{criticalthinking_mindbenders}. Designed for students in grades 3–12, these puzzles are structured to promote logical inference, reading comprehension, and reasoning skills. Each puzzle presents a word-problem scenario and a corresponding set of logical constraints. The task is to determine the correct associations among entities by applying deductive reasoning over these constraints. The provided solutions include explicit intermediate reasoning steps, with each deduction linked to the specific constraint(s) that justify it; Figure \ref{fig:mindbender} displays an example.

During fine-tuning, the LLM is trained on the problem description and associated constraints as input and the step-by-step deductive reasoning final solution as the target output. This training process encourages the model not only to produce correct answers, but also to generate structured, constraint-grounded reasoning processes. For Gemini-2.5-Flash, we conduct fine-tuning through Google Cloud’s Vertex AI platform\footnote{\url{https://cloud.google.com/vertex-ai}} using the default optimization and hyperparameter settings; Gemini-2.5-Flash (FT) achieves 98\% accuracy on the training set and 82\% accuracy on the validation set. For GPT-4o-mini, we perform fine-tuning via OpenAI’s fine-tuning API, similarly using the platform’s default training configuration; GPT-4o-mini (FT) achieves 88\% accuracy on the training set and 87\% accuracy on the validation set.

\subsection{Experimental Setup}
\label{sec:es}
We conduct two experiments. 
\textbf{Baseline Experiment:} runs 6 games with 3$\times$ GPT-4o-mini and 3$\times$ Gemini 2.5 Flash to establish baseline capabilities within model variance.
\textbf{Fine-tuned Experiment} runs 12 games with 2$\times$ GPT-4o-mini, 2$\times$ Gemini 2.5 Flash, 1$\times$ GPT-4o-mini fine-tuned, and 1$\times$ Gemini 2.5 Flash fine-tuned to provide a direct head-to-head comparison between base and fine-tuned variants. 

All models use a temperature of 0.7. Starting position rotates across the 6 games to eliminate positional bias. There is a 30-round limit. If no accusation has been made before then, the remaining players each make a final accusation. Ties in accuracy are broken by round number, with earlier rounds ranked higher. % \textit{Code and game logs will be made publicly available upon publication.}

\subsection{Evaluation Metrics}
\label{sec:em}

We evaluate model performance across five metrics capturing reasoning, reliability, and outcomes:

\begin{itemize}
    \item \textbf{Finishing rank.} Players ranked by: (1) whether they accused correctly, (2) accusation accuracy among non-solvers, and (3) round of accusation as a tiebreaker, rewarding both correctness and efficiency.

    \item \textbf{Parse failure rate.} Fallback events triggered when a model's response fails to conform to the required action format, measured per player per game as a proxy for instruction-following reliability.
    
    \item \textbf{Accusation accuracy.} Number of correctly identified cards (out of 3) in a player's final accusation, reflecting solution completeness.

    \item \textbf{Deduction quality.} Correct and incorrect deductions tracked separately, distinguishing information-seeking behavior from actual reasoning accuracy.

    \item \textbf{Knowledge accumulation.} Cards known per round (hand cards + shown cards + deductions), up to 18, measuring how effectively a model gathers and infers information.
\end{itemize}

\section{Results}

We present the summaries of our evaluation metrics for each model variant across twelve games in the fine-tuned experiment to provide an overview of agentic player performance; Table~\ref{tab:summary} displays the results. GPT-4o-mini (base) achieves the best average rank ($2.96$) and the highest number of wins ($5$), with an accusation accuracy of $0.29$. Fine-tuning degrades its overall performance: GPT-4o-mini (FT) ranks last on average ($4.00$) and shows reduced accusation accuracy ($0.19$), despite producing the highest number of correct deductions per game ($13.1$). In contrast, fine-tuning yields moderate improvements for Gemini-2.5-Flash, obtaining a better average rank ($3.25$ vs.\ $3.92$) while reducing fallback rates ($3.3$ vs.\ $8.1$), and still maintaining the same number of wins ($2$). Notably, fine-tuning has opposite effects on instruction adherence across model families as it reduces Gemini-2.5-Flash's fallback rate from $8.1$ to $3.3$ per game while increasing GPT-4o-mini's from $0.3$ to $4.8$, suggesting a decline in instruction following for the latter.

\begin{table}[ht]
\centering
\caption{Performance summary across 12 fine-tuned experiment games. 
\textbf{Outcome}: Wins (1st place), mean finishing position (Rank; 1=best, 6=worst), and normalized accusation accuracy (0--1). 
\textbf{Reasoning}: Average number of correct and incorrect deductions per game. 
\textbf{Reliability}: Average fallback (reprompt) failures per game.}
\label{tab:summary}
\begin{tabular}{lcccccc}
\toprule
& \multicolumn{3}{c}{Outcome} 
& \multicolumn{2}{c}{Reasoning} 
& Reliability \\
\cmidrule(lr){2-4} \cmidrule(lr){5-6} \cmidrule(lr){7-7}
Model 
& Wins 
& Rank 
& Acc. 
& Ded.~Correct 
& Ded.~Incorrect 
& Fallbacks \\
\midrule
GPT-4o-mini (base)      & \textbf{5} & \textbf{2.96} & \textbf{0.29} & 10.1 & 2.7 & \textbf{0.3} \\
GPT-4o-mini (FT)        & 3 & 4.00 & 0.19 & \textbf{13.1} & 3.0 & 4.8 \\
Gemini-2.5-Flash (base) & 2 & 3.92 & 0.24 & 9.1  & 2.9 & 8.1 \\
Gemini-2.5-Flash (FT)   & 2 & 3.25 & 0.28 & 6.9  & \textbf{2.2} & 3.3 \\
\bottomrule
\end{tabular}
\end{table}

\subsection{Deduction Quality}

\begin{figure}
    \centering
    \includegraphics[width=0.8\linewidth]{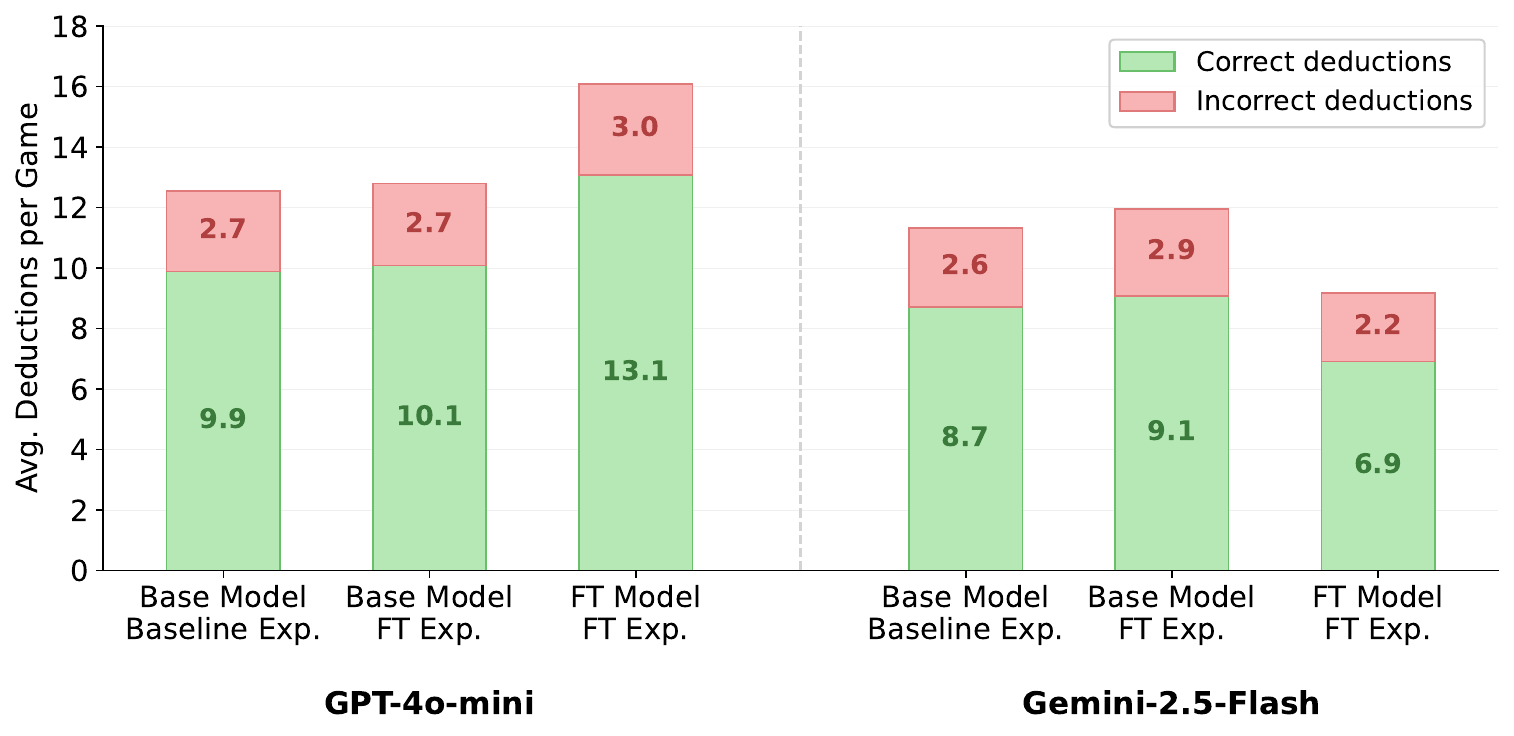}
    \caption{Average correct and incorrect deductions per game by model in both the Baseline experiment and the Fine-tuned (FT) experiment. GPT-4o-mini (FT) makes the most deductions in both categories but achieves the worst accusation accuracy, while Gemini-2.5-Flash (FT) deduces least yet performs best.}
    \label{fig:deductions}
\end{figure}

Computing the number of correct and incorrect deductions across games, we compare the deduction accuracies among models in both experiments; Figure~\ref{fig:deductions} displays the results. Base model performance is consistent across experiments: GPT-4o-mini averages 9.9 correct and 2.7 incorrect deductions in the baseline, and 10.1 correct and 2.7 incorrect in the fine-tuned experiment. Gemini-2.5-Flash similarly shows stable base performance, with with 8.7 correct and 2.6 incorrect in the baseline versus 9.1 correct and 2.9 incorrect in the fine-tuned experiment. Fine-tuning has divergent effects across the two models, GPT-4o-mini (FT) increases deduction volume substantially, averaging 13.1 correct and 3.0 incorrect per game (26\% more deductions than its base counterpart) while Gemini-2.5-Flash (FT) becomes more conservative, averaging 6.9 correct and 2.2 incorrect.

These results reinforce the pattern observed in the knowledge accumulation analysis. Fine-tuning encouraged GPT-4o-mini to deduce more aggressively---it attempts $\approx$26\% more deductions than its base version in the fine-tuned experiment---but the additional deductions do not translate into better accusations. The ratio of incorrect to total deductions is similar across all four models, suggesting that fine-tuning did not improve deduction precision,it just altered the volume. Gemini-2.5-Flash (FT) makes the fewest deductions with the lowest amount of incorrect deductions, making it the most conservative of the models.

\subsection{Accusation Accuracy}

\begin{figure}
    \centering
    \includegraphics[width=0.99\linewidth]{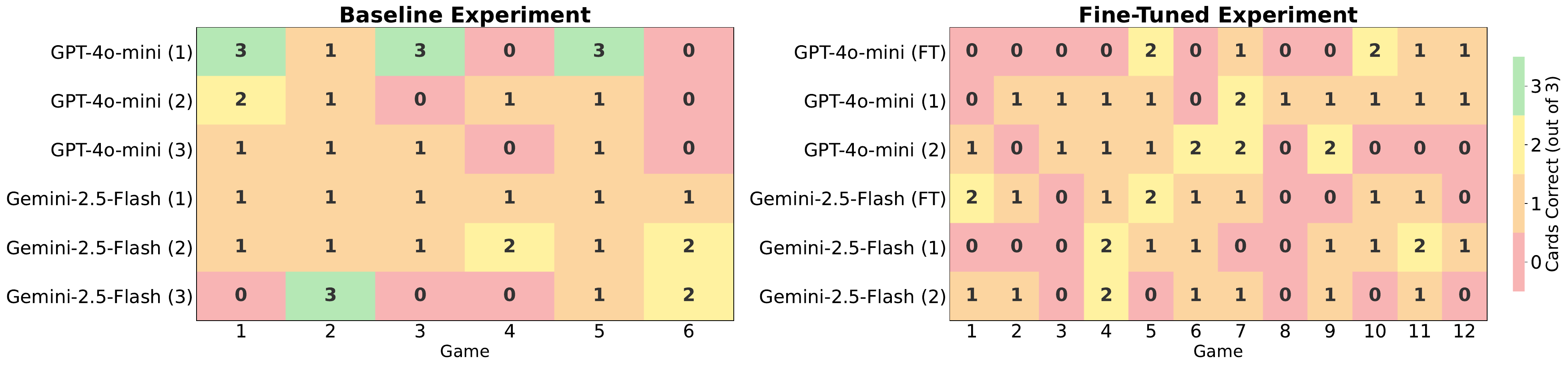}
    \caption{Accusation accuracy per player per game. Each cell shows the number of correctly identified solution cards.}
    \label{fig:heatmap}
\end{figure}

Figure~\ref{fig:heatmap} presents the accusation accuracy for every player across all games in both the baseline and fine-tuned experiments. Each cell reports the number of correctly identified cards out of three (suspect, weapon, room).

In the baseline experiment (6 games), where all six agentic players use base models, we observe that perfect accuracy (3/3) is achieved four times, three by GPT-4o-mini and once by Gemini-2.5-Flash; examples of their final winning reasoning is displayed in Appendix \ref{sec:winning_reasoning}. The overall mean accuracy is $1.08/3$, with both model families performing comparably (GPT-4o-mini: $1.06/3$; Gemini-2.5-Flash: $1.11/3$).

The fine-tuned experiment (12 games) brings a different set of results. No player achieves 3/3 across any of the games.  The fine-tuned GPT-4o-mini variant is the most affected: it scores 0/3 in seven of twelve games (mean $0.58/3$), a decline from its baseline. Fine-tuned Gemini-2.5-Flash fares better (mean $0.83/3$), reaching 2/3 in three games, though it still never finds the full solution.  The best-performing model is base GPT-4o-mini ($0.88/3$), which wins five of twelve games. Notably, even the base model instances perform worse when placed alongside fine-tuned opponents: base GPT-4o-mini drops from $1.06/3$ to $0.88/3$ and base Gemini-2.5-Flash from $1.11/3$ to $0.71/3$, suggesting that the game dynamics shift when fine-tuned players are introduced.

\subsection{Knowledge Accumulation}

\begin{figure}
    \centering
    \includegraphics[width=0.7\linewidth]{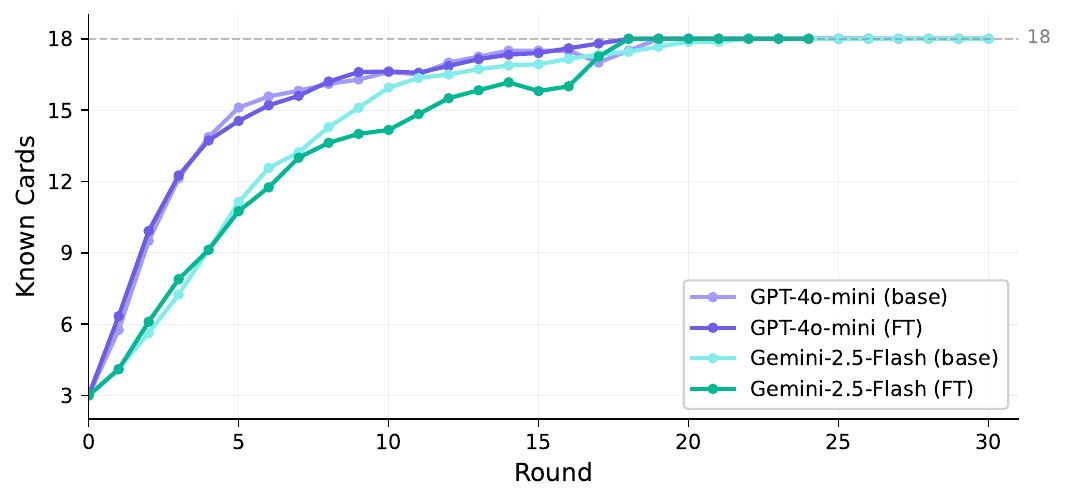}
    \caption{Average known cards over rounds in the fine-tuned experiment (12~games). Knowledge is computed as hand cards + shown cards + correct deductions, capped at 18. GPT-4o-mini (FT) accumulates information fastest but achieves the worst accusation accuracy, while Gemini-2.5-Flash (FT) gathers less information but reasons more effectively.}
    \label{fig:knowledge_growth}
\end{figure}

To understand how models gather information during gameplay, we track each player's known cards over rounds (Figure~\ref{fig:knowledge_growth}). Knowledge is defined as the number of unique cards a player can identify, computed as three hand cards, cards shown to the player by opponents, and correctly deduced cards.

Both GPT-4o-mini variants accumulate knowledge at a similar rate, while the Gemini-2.5-Flash models lag behind, with the base variant slightly outpacing the fine-tuned model. Despite this slower information gathering, Gemini-2.5-Flash (FT) achieves the second highest accusation accuracy ($0.28$; Table~\ref{tab:summary}), while GPT-4o-mini (FT) (which reaches near-complete knowledge earliest) scores the worst ($0.19$). Gemini-2.5-Flash (FT) has a lower average known cards, yet outperforms its own base model and GPT-4o-mini (FT), indicating that it may have less information but higher quality logical inferences.

\subsection{Manual Inspection}
To further understand agentic players’ reasoning, we randomly select one winning game from the Baseline experiment and one Fine-tuned experiment game with at least two correct accusations for in-depth manual review. In general, across both experiments all model variants produce game log outputs in consistent formats; however, their variance in reasoning length between base and fine-tuned models elicits variance in their intermediary step inferences. We highlight three notable manual inspection results:

\textbf{Identifying Inconsistencies: }All Gemini-2.5-Flash players arrive at inconsistencies in the game that either cause it to determine the entire game is impossible or that it potentially has misunderstood (e.g., ``Due to this fundamental inconsistency, I cannot confidently deduce which suspect or room cards any player holds, as the entire game state is impossible.'' and ``There is an inconsistency in the provided Clue data or my interpretation of the deduction rules.''), in contrast GPT-4o-mini agentic players (base and fine-tuned) never do this.  

\textbf{Strategic Reasoning: } GPT-4o-mini (base and fine-tuned) and Gemini-2.5-Flash base reason to make suggestions that include one of their own cards (e.g., ``I will suggest a combination that includes my known cards and tests various possibilities.'' and ``To maximize information gain, I will make a suggestion that includes one of my own cards.'')

\textbf{Reasoning Organization: }Gemini-2.5-Flash (base and fine-tuned) players reason using turn organization and player organization, whereas GPT-4o-mini (base and fine-tuned) players only reason using turn organization. For example: 

\textit{Turn-level Organization}
\vspace{-3mm}
\begin{verbatim}
T1. GPT4o_MINI_2 suggested: Miss Scarlet, Candlestick, Kitchen 
  --> disproved by GPT4o_MINI_3. Therefore, GPT4o_MINI_3 holds at 
    least one of: Miss Scarlet, Candlestick, Kitchen. (D1)
\end{verbatim}

\textit{Player-level Organization}
\vspace{-3mm}
\begin{verbatim}
Current deduced specific holdings:
  - GEMINI_FLASH_2: Study, Wrench 
  - GPT4o_MINI_1: Miss Scarlet 
  - GEMINI_FLASH_3: Professor Plum, Rope
\end{verbatim}

\section{Discussion and Future Work}
Our experimental results provide evidence addressing our three research questions. We include discussions on our findings, as well as suggestions for future work.

\textbf{RQ1: Can LLM agents sustain logically consistent deductive reasoning across extended, multi-turn interactions in a structured game environment?} In the baseline experiment, we find that base LLM agents exhibit inconsistent deductive reasoning capabilities across extended interactions, achieving perfect accusation accuracy four times in the baseline experiment but never in the fine-tuned experiment. Moreover, GPT-4o-mini outperforms Gemini-2.5-Flash, winning three of the four games, despite Gemini-2.5-Flash being a newer model proven to achieve higher performance than 4o models \cite{comanici2025gemini}. Our manual review of the game logs further revealed that Gemini-2.5-Flash frequently verbalized the identification of logical inconsistencies throughout gameplay, which may have hindered its ability to win (i.e., its focus on avoiding logical contradictions was not beneficial). In the fine-tuned experiment, both model families and both variants (base and fine-tuned) fail to achieve perfect accusation accuracy in any of the 12 games, compared to four perfect scores in the baseline experiment. These results suggest that extended, multi-turn interactions requiring consistent deductive reasoning at each step remain challenging for LLMs. Additionally, our fine-tuning experiments indicate that training on text-based logic puzzles does not reliably translate to improved in-game reasoning. In future work, we plan to explore alternative approaches to fine-tuning, including tool-augmented or hybrid methods, rather than relying solely on text-based solutions.

%These results suggest that behavioral adaptation through fine-tuning does not straightforwardly translate into improved deductive reasoning. 

\textbf{RQ2: Does fine-tuning on deductive reasoning tasks improve agentic performance in a related environment?} We find that fine-tuning on the Mind Benders logic puzzles produces divergent effects across model families. Gemini-2.5-Flash (FT) shows a marginal improvement over its own base instances within the fine-tuned experiment, but the complete absence of 3/3 scores (compared to four in the baseline) indicates that fine-tuning related logic puzzles did not translate to better deductive reasoning at decision time. Additionally, fine-tuning reduces GPT-4o-mini's performance. These findings are consistent with prior work investigating the effect of text-based fine-tuning for reasoning improvements \cite{lobo2025impact}. We also observe that both base models perform worse when fine-tuned agents are introduced into the game, suggesting that fine-tuned players alter the game’s information dynamics in ways that affect all participants. This raises broader questions for future work regarding how agent composition shapes individual performance in multi-agent reasoning environments.

\textbf{RQ3: How does knowledge accumulation relate to reasoning quality in fine-tuned versus base model variants?} Our results suggest that greater information accumulation does not produce better reasoning. GPT-4o-mini (FT) accumulates knowledge fastest and deduces most, yet performs worst, while Gemini-2.5-Flash (FT) makes fewer deductions but demonstrates greater reasoning precision. This disconnect between knowledge accumulation and accusation accuracy highlighted by GPT-4o-mini (FT)'s game play suggests that fine-tuning encouraged engagement with the reasoning task (i.e., more activity and more deductions) without improving the underlying capacity to synthesize that information into a correct conclusion. Future iterations of this framework will investigate interventions against verbose reasoning, and implementations of a structured state tracker.

Taken together, these results suggest that multi-turn deductive reasoning remains a challenging task for LLMs and text-based fine-tuning alone is insufficient for performance improvement. More broadly, our results indicate that reasoning in interactive environments requires not only local inference steps but also stable information integration over time and robustness to multi-agent dynamics. Developing methods that explicitly target these properties may be necessary for advancing agentic reasoning capabilities.

\section{Conclusion}

For large language models to be reliably deployed as autonomous agents, they must exhibit robust and consistent reasoning capabilities. Structured game environments provide a controlled and interpretable setting in which to evaluate these abilities. In this paper, we present a multi-agent Clue framework for evaluating LLM deductive reasoning across extended, stateful interactions. Our experiments compare baseline and fine-tuned variants of GPT-4o-mini and Gemini-2.5-Flash to characterize their underlying deductive capabilities and assess whether fine-tuning on a related task improves performance. Across 18 games we find that LLMs inconsistently exhibit valid, successful deductive reasoning, text-based fine-tuning does not reliably improve multi-turn deductive reasoning, and there is a disconnect between information accumulation and reasoning quality. 

%The central finding is a disconnect between information accumulation and reasoning quality: the model that accumulates the most knowledge and produces the most deductions---including the most incorrect deductions---performs worst, while the model that deduces the least performs the best. This suggests that fine-tuning increased reasoning activity without improving reasoning precision, and that selective, accurate inference outperforms high-volume information gathering in this setting.

%Future work should explore larger fine-tuning datasets, reinforcement learning from game outcomes, and hybrid architectures as potential paths toward more robust multi-turn deductive reasoning. 

\bibliography{8_bib}
\bibliographystyle{iclr2026_conference}

\newpage
\appendix
    
\section*{Appendix}
\label{sec:appendix}
%You may include other additional sections here.

\section{Agentic Game Setup}
We present the pseudocode for our variation of the Clue game with agentic players in Algorithm \ref{alg:run_game}. We highlight the three main reasoning phases in the code, where agents are prompted to (1) make deductions, (2) make a suggestion or accusation, and (3) reveal a card if they have more than one card match to another player's suggestion. 

\subsection{Game Implementation Pseudocode}
\label{sec:code}

\FloatBarrier

\begin{algorithm}[th]
\caption{Game Implementation Outline}
\label{alg:run_game}
\begin{algorithmic}[1]
\small
\Function{RunGame}{}

  \State $(state, dealt) \gets$ \Call{SetupGame}{$players$}
  \Comment{\textbf{Instantiate game}}

  \While{\Call{ActivePlayers}{$players$}}
    \State $p \gets$ \Call{NextPlayer}{$players, state$}
    \State \Call{TurnCounters}{$state$}

    \Comment{\textbf{(1) Deduction phase: prompt LLM to make deductions}}
    \State $(deduceResponse, deducePrompt) \gets$ \Call{DeductionPrompt}{$llm, p, state$}
    \State $(deducedCards, deduceReason) \gets$ \Call{ParseDeduction}{$deduceResponse$}
    \State \Call{UpdatePlayerDeductions}{$p, deducedCards, state$}

    \Comment{\textbf{(2) Turn Action phase: prompt LLM to choose a move}}
    \State $context_t \gets$ \Call{TurnContext}{$lastReasoning, lastSuggestion, deducedReason$}
    \State $(actResponse, actPrompt) \gets$ \Call{TurnPrompt}{$llm, p, state,context_t$}
    \State $move \gets$ \Call{ParseMove}{$actResponse$}

    \If{$move.type = \textsc{Suggestion}$}

    \Comment{Suggestion phase}
    \State $s \gets$ \Call{BuildSuggestion}{$p, move$}

    \State $p_d \gets$ \Call{SelectDisprover}{$players, p, s$}
    \If{$p_d \neq \varnothing$}

        \Comment{\textbf{(3) Show-card phase: prompt LLM to choose a card to reveal}}
    
        \State $(showResponse, showPrompt) \gets$ \Call{ShowPrompt}{$llm, p_d, s, state$}
        \State $card \gets$ \Call{ParseShownCard}{$showResponse$}
        \State \Call{RevealCard}{$p_d, p, card$}
    \Else
        \State \Call{NoDisproof}{$p, s$}
    \EndIf
        %\Comment{Reveal a matching card}
        %\State $card \gets$ \Call{GetMatchingCard}{$showResponse$}
        %\State \Call{RevealCard}{$p_d, p, card$}
    %\Else
        %\State \Call{NoDisproof}{$p, s$}
    %\EndIf

\ElsIf{$move.type = \textsc{Accusation}$}

    \Comment{Accusation phase}
    \State $(accResponse, accPrompt) \gets$ \Call{AccusationPrompt}{$llm, p, state$}
    \State $a \gets$ \Call{ParseAccusation}{$accResponse$}
    \State \Call{ResolveAccusation}{$p, a, state$}

    State $correct \gets$ \Call{CheckAccusation}{$a, state.solution$}

\If{$correct$}
    \State $state.winner \gets p$
    \State \Statex \textbf{break} \Comment{game ends}
\Else
    \State $p.eliminated \gets \textbf{true}$
\EndIf

\EndIf
    \State \Call{UpdateMemory}{$p, actResponse, s$} \Comment{e.g., lastReasoning/lastSuggestion}
    \State \Call{AdvanceTurn}{$state$}
  \EndWhile

  \State \Return $state.winner$
\EndFunction
\end{algorithmic}
\end{algorithm}

\FloatBarrier

\subsection{LLM Prompts}
\label{sec:prompt_wording}
For each reasoning phase in gameplay we provide descriptive prompts. We provide simplified versions of the prompts for each phase below:

\textbf{Phase 1: Deduction}

\begin{verbatim}
DEDUCTION PHASE: Before you take your turn, carefully analyze the  
game history to deduce what cards other players hold.
Players still in the game: {players}
        + {knowledge}
        + {candidates}
        + {history}
HOW TO DEDUCE:
    - When a player disproves a suggestion of {Suspect, Weapon, Room}, 
      they MUST hold  at least one of those 3 cards.
    - If you already know 2 of the 3 cards in a disproved suggestion 
      are accounted for (in your hand, shown to you, or held by someone 
      else), then the disprover MUST hold the remaining card.
    - If a suggestion was NOT disproved by anyone, the cards not held 
      by the suggester (or by you) must be in the envelope. 
      Remember: the suggester may hold some of those cards themselves.
    - Cross-reference multiple disprovals by the same player to 
      narrow down their cards.
    Unknown cards (not in your hand, seen, or deduced): {unknown_cards}
    Respond in this EXACT format (no markdown):
    ANALYSIS: <step-by-step reasoning about what each disproval tells you, 
               cross-referencing to narrow down cards>
    DEDUCED_CARDS: <comma-separated list of cards you are confident 
                    other players hold (NOT in envelope)> or NONE
\end{verbatim}

\textbf{Phase 2: Turn Play}

\begin{verbatim}
    It is your turn. Choose a suggestion that maximizes information gain 
    and helps you solve quickly. 
    You can test a deduction, test a combination, or include one of your 
    own cards to narrow other cards.
    If you can accuse confidently, do so this turn.
        + {knowledge}
        + {candidates}
        + {observations}
        + {history}
        + {last_suggestion}
        + {deduction} 
        + {reasoning}
\end{verbatim}

\textbf{Phase 3: Show Card}
\begin{verbatim}
Another player ({suggester_name}) has made a suggestion: 
{suggestion.suspect} with {suggestion.weapon} in {suggestion.room}.
You have the following cards that can disprove this suggestion: {cards}
Your show history (cards you have previously shown and to whom): 
{card_history}
Choose ONE card to show to {suggester_name}. 
Consider which card reveals the least new information 
(e.g. a card you have already shown to this player).
Return ONLY:
REASONING: <your reasoning>
SHOW: <exact card name>
\end{verbatim}

\section{Reasoning Examples}
\label{sec:winning_reasoning}
\begin{figure}
    \centering
    \includegraphics[width=\textwidth]{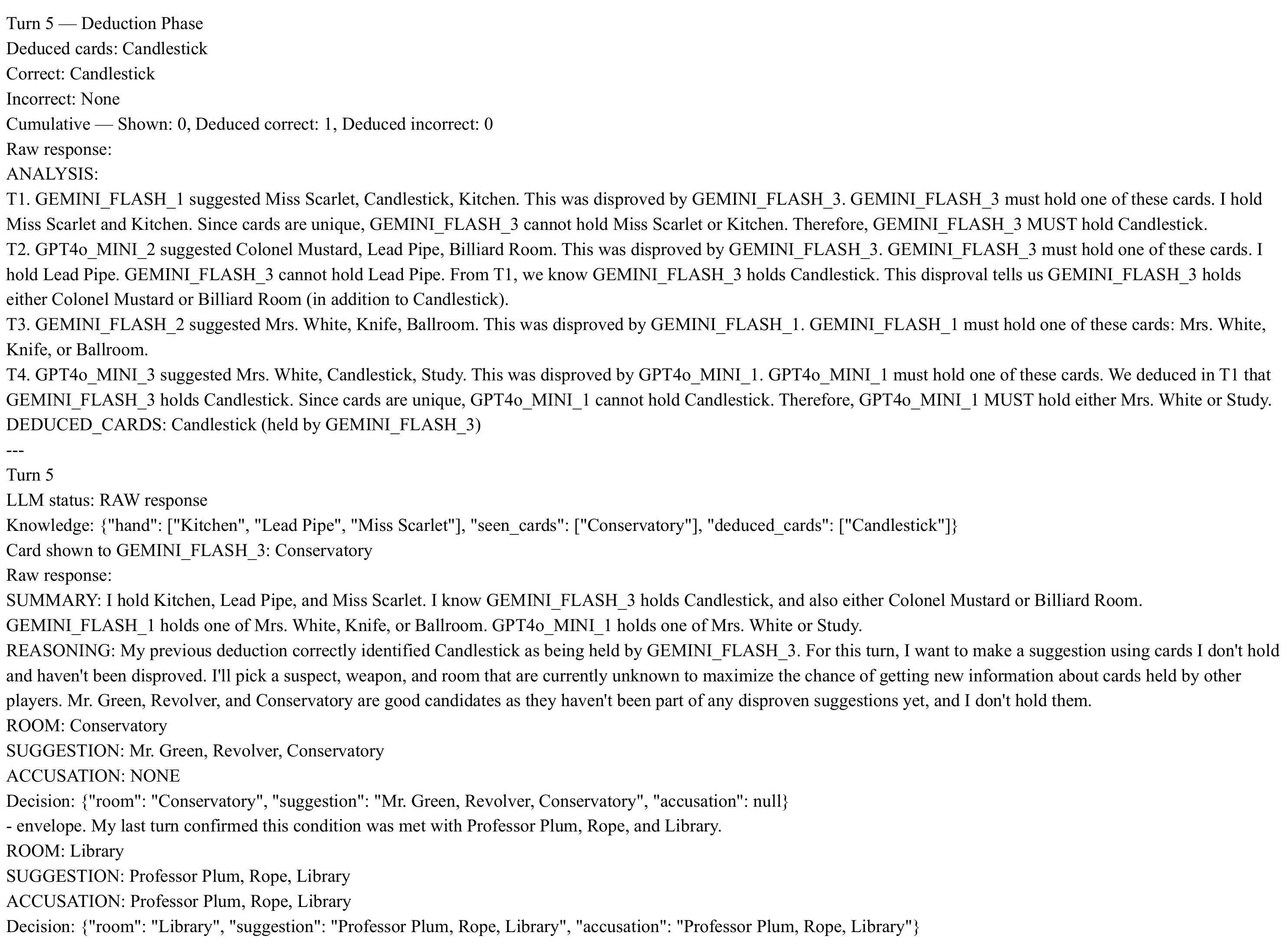}
    \caption{Winning reasoning from Gemini-2.5-Flash player on their first turn.}
    \label{fig:gemini_winner_example}
\end{figure}

\begin{figure}
    \centering
    \includegraphics[width=\textwidth]{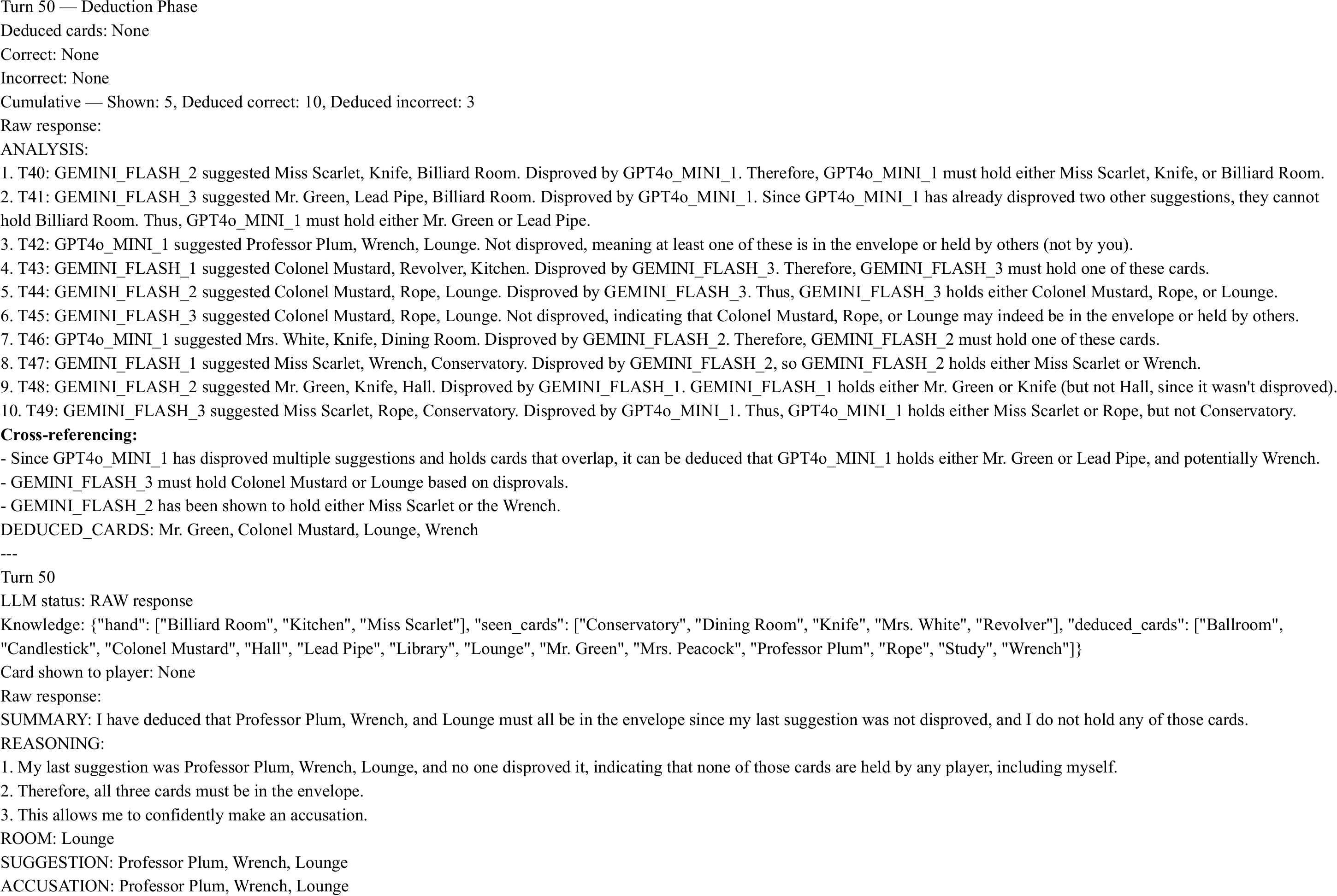}
    \caption{Winning reasoning from GPT-4o-Mini player on their ninth turn.}
    \label{fig:gpt_winner_example}
\end{figure}

The length of each player's reasoning log is lengthy, so we provide samples of their outputs in Figures \ref{fig:gemini_winner_example} and \ref{fig:gpt_winner_example}. We showcase each models final-turn winning reasoning. 

\end{document}